\documentclass[conference]{IEEEtran}
\IEEEoverridecommandlockouts
\usepackage{amsmath,amssymb,amsfonts}
\usepackage{algorithmic}
\usepackage{graphicx}
\usepackage{textcomp}
\usepackage{xcolor}
\usepackage{tabularx,booktabs}
\usepackage[hidelinks]{hyperref}
\usepackage{subcaption}
\usepackage[
    backend=biber,
    style=ieee,
    isbn=false,
    url=false,
    eprint=false
]{biblatex} 
\graphicspath{ {./figures/} }

\def\BibTeX{{\rm B\kern-.05em{\sc i\kern-.025em b}\kern-.08em
    T\kern-.1667em\lower.7ex\hbox{E}\kern-.125emX}}

\begin{document}

\title{Robotic Learning in your Backyard: A Neural Simulator from Open Source Components
\thanks{This work was partly supported by the Engineering and Physical Sciences Research Council Grant [EP/S023917/1]}
}

\author{
    \IEEEauthorblockN{
        Liyou Zhou\IEEEauthorrefmark{1}\IEEEauthorrefmark{2}\textsuperscript{[0009-0005-9491-9003]},
        Oleg Sinavski\IEEEauthorrefmark{3}\textsuperscript{[0009-0002-8529-8100]},
        Athanasios Polydoros\IEEEauthorrefmark{2}\textsuperscript{[0000-0002-4597-0567]}
    }
    \IEEEauthorblockA{
        \textit{\IEEEauthorrefmark{1} Department of Engineering,
        University of Cambridge,
        Cambridge, UK}
    }
    \IEEEauthorblockA{
        \textit{\IEEEauthorrefmark{2} Lincoln Centre for Autonomous Systems (L-CAS),
        University of Lincoln,
        Lincoln, UK}
    }
    \IEEEauthorblockA{
        \textit{\IEEEauthorrefmark{3} Wayve Technologies Ltd,
        London, UK}
    }
}

\maketitle

\begin{abstract}
The emergence of 3D Gaussian Splatting for fast and high-quality novel view synthesize has opened up the possibility to construct photo-realistic simulations from video for robotic reinforcement learning. While the approach has been demonstrated in several research papers, the software tools used to build such a simulator remain unavailable or proprietary. We present SplatGym, an open source neural simulator for training data-driven robotic control policies. The simulator creates a photorealistic virtual environment from a single video. It supports ego camera view generation, collision detection, and virtual object in-painting. We demonstrate training several visual navigation policies via reinforcement learning. SplatGym represents a notable first step towards an open-source general-purpose neural environment for robotic learning. It broadens the range of applications that can effectively utilise reinforcement learning by providing convenient and unrestricted tooling, and by eliminating the need for the manual development of conventional 3D environments.
\end{abstract}


\section{Introduction}

In recent years, deep reinforcement learning (DRL) has been increasingly used to train robot control policies \cite{moralesSurveyDeepLearning2021}. DRL is a model-free approach  in which the control policy is learned directly from raw vision inputs. The policy is usually represented by a neural network and trained using reinforcement learning algorithms. Termed Vision Action Models (VAMs), this approach has the distinct advantage of being adaptable to new environments and tasks. Furthermore, the policy can be trained in simulation and transferred zero-shot to the real world. This is particularly useful in applications, where the environment is complex and the cost of in situ training is high. This approach first demonstrated its potential in the Autonomous Driving domain (\cite{kendallLearningDriveDay2019}) and has been further applied to a wide range of robotic tasks with promising results (\cite{octomodelteamOctoOpenSourceGeneralist2024, VisualCortexVC1, kimOpenVLAOpenSourceVisionLanguageAction2024, haarnojaLearningAgileSoccer2023, heLearningRepresentationsThat2022}).

The Conventional 3D graphics environments used in training these policies require manual creation of object models and textures \cite{MuJoCoAdvancedPhysics} \cite{BulletRealTimePhysics2022}. This has a few disadvantages:
\begin{enumerate}
    \itemsep0em
    \item Require very specialised expertise to build models and textures. Hence, it is time-consuming and cost-prohibitive.
    \item The visual appearance of the environment diverge between the real world and the simulation, leading to poor sim-to-real transferability of the policy.
\end{enumerate}

The field of novel view generation has seen rapid advances in recent years. Using NeRF/Gaussian Splatting techniques, novel view images can be generated from a few input images of a scene (\cite{liuInstanceNeuralRadiance2023,kerbl3DGaussianSplatting2023}). It provides an alternative approach to building 3D simulators. The simulator can be built directly from videos taken from the real world and can generate photo-realistic images of the scene from arbitrary camera view points. In \cite{haarnojaLearningAgileSoccer2023}, a soccer-playing robot is trained in NeRF simulation and \cite{byravanNeRF2RealSim2realTransfer2022} used NeRF to train a robot to navigate autonomously in an indoor environment. Both reported a diverse range of emergent locomotion skills and good sim-to-real transferability.

\begin{figure}[!t]
    \centering
    \includegraphics[width=0.49\textwidth]{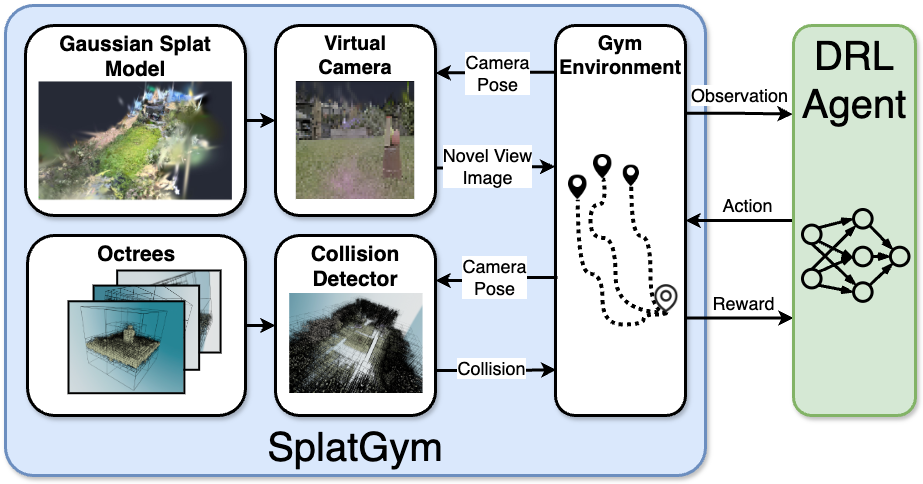}
    \caption{Overview of SplatGym, an open-source neural simulator for training data-driven robot control policies. The simulator combines novel view generation and fast collision detection to create a photorealistic simulation environment for reinforcement learning.}
    \label{fig:splatgym_simulator_overview.drawio}
\end{figure}

While the DRL with NeRF simulator approach is promising, software tools for constructing such a simulator remain unavailable or proprietary. Documentation and guidance is sparse. Not only are the aforementioned research papers irreproducible, but this issue has also become a significant barrier to advancing research in DRL. Therefore, in this paper, we introduce SplatGym, an open source NeRF/Gaussian Splatting based simulator for training data-driven robot control policies (Fig~\ref{fig:splatgym_simulator_overview.drawio}). The simulator is built on top of existing open-source software and fills a gap in DRL research tooling. The primary contributions of this paper are:
\begin{enumerate}
    \item An ego camera simulator for vision-action policy learning. Real-time speed is achieved on a consumer-grade GPU via the use of Gaussian Splatting.
    \item A pipeline for fast and efficient collision detection using octree representation.
    \item A modular software stack that integrates with popular reinforcement learning libraries.
    \item The project is fully open source and available at \url{github.com/SplatLearn/SplatGym}
\end{enumerate}

\section{Related Work}

\subsection{Visual Navigation}

Humans can navigate in unseen environments effortlessly using only vision inputs. The task of visual navigation is to train a robot to navigate to a goal position using only vision inputs (\cite{zhangSurveyVisualNavigation2022}). Existing approaches generally fall into 2 categories:

\begin{enumerate}
    \item Geometry-based methods: navigation is achieved through a pipelined approach of mapping, localization, planning.
    \item Embodied AI style: End to end neural
    network is trained in a virtual environment for direct control of locomotion from vision inputs.
\end{enumerate}

In a domestic scene, \cite{krantzNavigatingObjectsSpecified2023} employs a modular approach and split the task into exploration, goal instance re-identification, goal localization, and local navigation. It relies on an underlying occupancy map to navigate around the environment. In contrast, \cite{zhuTargetdrivenVisualNavigation2017} uses an end-to-end learning approach. The goal of the robot is to navigate towards a pre-defined target in as few steps as possible. The end-to-end model takes the ego observation image and the target image as input. Trained with reinforcement learning, it outputs an optimal action for the robot at each step. The robot navigates to different targets in the scene without re-training.

The geometry-based methods suffer from noise sensitivity and error accumulation. While the embodied AI style methods are data hungry. However, embodied AI is more flexible and can adapt to new environments quickly.

Using an embodied AI approach, the realism of the simulator becomes crucial. Because there is no separation of concern as in a pipelined system, any deviation in the visual appearance propagates directly to the output. A lot of research has been conducted on building realistic simulation environments for different applications. \cite{szotHabitatTrainingHome2022} focused on building a photo realistic 3D graphics simulator for home environments. The associated Habitat challenge (\cite{habitatchallenge2023}) tasks an agent to navigating to a location 'described' by an image when initialized in an unfamiliar scene. In \cite{zhuTargetdrivenVisualNavigation2017}, in order to evaluate the performance of the proposed model, a full-featured 3D simulator had to be built using a game engine.

\subsection{Neural Radiance Fields for Novel View Synthesis}

Proposed in \cite{mildenhallNeRFRepresentingScenes2020}, NeRF is a technique that can generate novel view images from a few input images of a scene. It models a scene as a radiance field and approximates it using a fully connected neural network. An accurate 3D scene representation can be learned through minimising a photogrammetry loss between ground truth and synthesized images. \cite{mullerInstantNeuralGraphics2022} drastically reduces the computation cost of NeRF. The paper reports rendering of similar quality to NeRF at 2 orders of magnitude faster speed.

\cite{kerbl3DGaussianSplatting2023}, further improves the speed and quality of NeRF by replacing the neural network with a 3D Gaussian mixture. The scene is represented by a comparatively low number of Gaussians centred around areas where surface features have been detected. This avoids computation in empty spaces and allows for more efficient training and inference. Real-time rendering and display have been reported to be possible with Gaussian splatting ($\geq 30$ fps at 1080p resolution) while matching the visual quality of radiance field representation.

\subsection{Neural Simulators for Robotics}

Since its inception, NeRF has quickly shown potential in training robot control policies by providing realistic simulated viewpoints.

\cite{tancikBlockNeRFScalableLarge2022} built a large-scale NeRF model of the streets of San Francisco. The simulator is then used to train autonomous driving models. \cite{shenDriveEnvNeRFExplorationNeRFBased2024} also targeted autonomous driving with a NeRF simulator. They used appearance embeddings in the NeRF model to simulate different weather conditions. The simulator is used to train a driving policy, which is shown to transfer well to the real world.

\cite{byravanNeRF2RealSim2realTransfer2022} used NeRF simulation to train a bipedal robot to navigate in an indoor scene. A marching-cubes algorithm is used to obtain the surface mesh of obstacles in the scene to perform collision detection in the simulator. The paper also explored overlaying interactive elements into the simulation. The robot is tasked with pushing a ball around a room. The ball is rendered using conventional 3D graphics and physical simulation, while the rest of the scene is rendered using NeRF. The rendering of the ball is simply overlaid on top of the NeRF-generated 2D image to create the final image. The follow-up work, \cite{haarnojaLearningAgileSoccer2023}, used this exact method to train a soccer-playing robot showing impressive emergent locomotion and manipulation skills from the RL process.

Despite the corpus of research published, the simulators used in these works are not available to the public or have significant licensing restrictions. Table.~\ref{tab:simulators_comparison} compares the features of the simulators and highlights how SplatGym fills the gap in research tooling.

\begin{table*}[!t]
    \centering
    \caption{Comparison of neural simulators used in recent robotics research.}
    \label{tab:simulators_comparison}
    \begin{tabular}{ccccccc}
        \toprule
        Simulator & Novel View Synthesis & Collision Detection & DRL Interface & Publicly Available & Open Source & Licence \\
        \midrule
        BlockNeRF \cite{tancikBlockNeRFScalableLarge2022} & NeRF & No & N/A & No & No & N/A \\
        DriveEnvNeRF \cite{shenDriveEnvNeRFExplorationNeRFBased2024} & NeRF & Yes (via Unity) & Nvidia Omniverse & Partial & Partial & Proprietary \\
        NeRF2Real \cite{byravanNeRF2RealSim2realTransfer2022} & NeRF & Yes (via MuJoCo) & N/A & No & No & N/A \\
        \textbf{SplatGym (Ours)} & Gaussian Splatting & Yes (via FCL) & Gymnasium \cite{towersGymnasium2024} & Yes & Yes & Apache 2.0 \\
        \bottomrule
    \end{tabular}
\end{table*}

\section{Methodology}

\subsection{Problem Statement}

The goal of the simulator is to enable the training of data-driven robot control policies. The robot, using only an ego camera image as an input, should be able to generate movement commands to navigate to pre-defined goal positions in the simulator. Building from a short video of a scene, the simulator should generate photo-realistic ego camera images for arbitrary robot poses. It should also detect collisions between the robot and the underlying scene to teach the robot obstacle avoidance. Therefore, there are 2 main components of the simulator:

\begin{enumerate}
    \item Novel View Synthesis - Generate photo-realistic images of the scene from any arbitrary camera pose.
    \item Collision Detection - Detect collision between the robot and the underlying scene objects.
\end{enumerate}

\subsection{Novel View Synthesis}

NeRF scenes can be reliably trained from a few images using the software package nerfstudio\cite{tancikNerfstudioModularFramework2023}. The package provides several algorithms for training and rendering. Of the methods provided, the Gaussian splatting (GS) algorithm exhibits many desirable properties suitable for the task of the simulator.

\begin{enumerate}
    \item GS is much faster than radiance field based methods in both training and rendering. Reinforcement learning requires numerous samples to train. A simulator must use computing resources efficiently.
    \item GS can better recreate details in close-up views, even in a large and complex scene (\cite{kerbl3DGaussianSplatting2023}). This is desirable in natural and unstructured scenes.
\end{enumerate}

\begin{figure}
    \centering
    \includegraphics[width=0.24\textwidth]{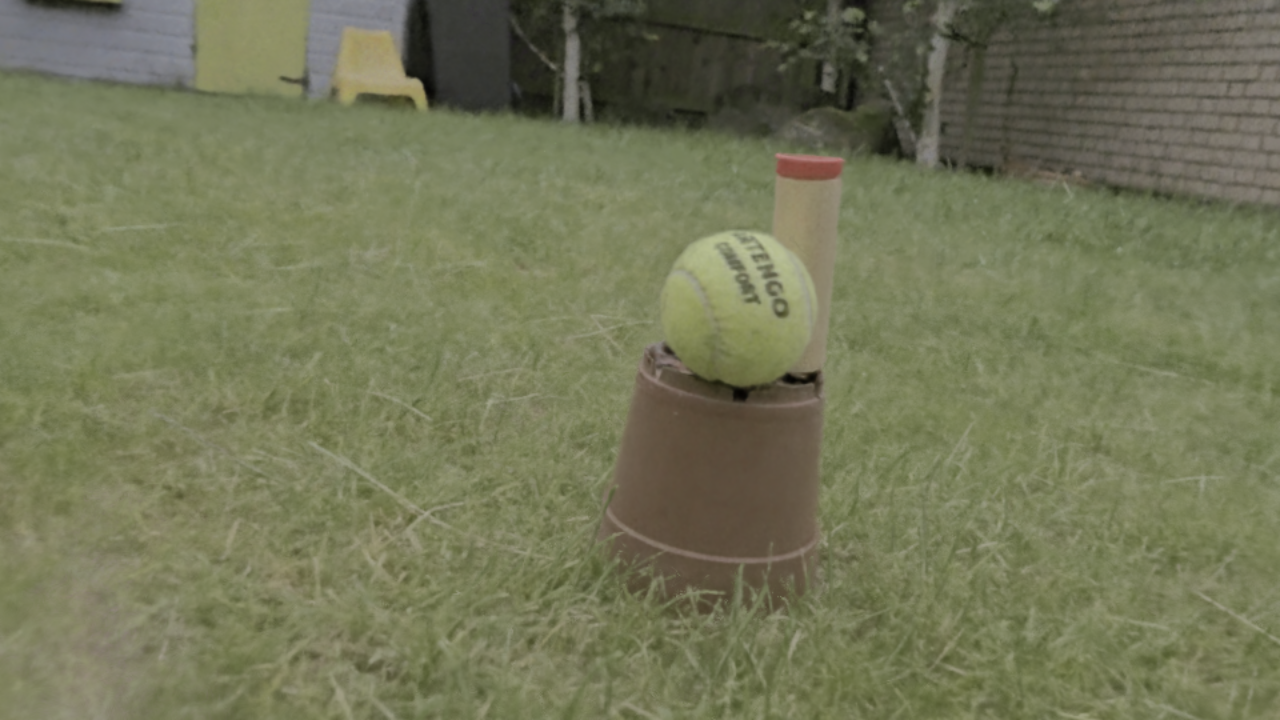}
    \includegraphics[width=0.24\textwidth]{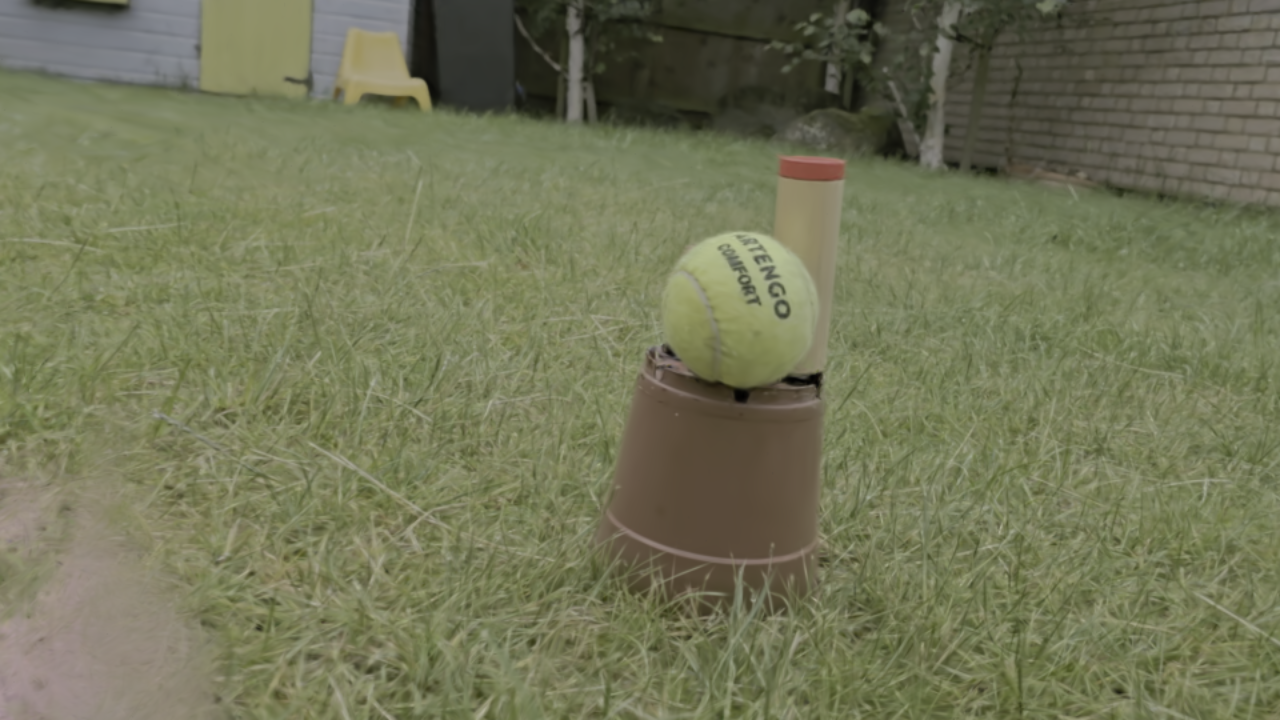}
    \caption{Novel view images of the same scene and same pose, trained and rendered by NeRF (left) and Gaussian Splatting (right).}
    \label{fig:splatfacto_vs_nerfacto}
\end{figure}

Splatfacto, the Gaussian Splatting algorithm implemented in nerfstudio, enhances the algorithm presented in \cite{kerbl3DGaussianSplatting2023} with existing features in nerfstudio. Training from the same data, a comparison is done against nerfacto, the corresponding NeRF implementation. The result (Tab.\ref{tab:splatfacto_vs_nerfacto}) shows that splatfacto is 45\% faster in training and 291 times faster in inference. The quality of the rendered images is comparable as shown in Fig.~\ref{fig:splatfacto_vs_nerfacto}. splatfacto is hence used for the simulator.

\begin{table}[h]
    \centering
    \caption{Speed comparison between nerfacto and splatfacto}
    \label{tab:splatfacto_vs_nerfacto}
    \begin{tabular}{ccc}
        \toprule
        \textbf{Algorithm} & \textbf{Training Time} & \textbf{Inference Time} \\
        \midrule
        & ($ms/iteration$) & ($ms/frame$) \\
        \midrule
        nerfacto & $31$ & $2270$ \\
        splatfacto & $17$ & $7.80$ \\
        \bottomrule
    \end{tabular}
\end{table}

Using nerfstudio and its associated tooling, videos can be processed to extract key frames camera poses, and the Gaussian Splatting model can be trained and saved. The model can be loaded at runtime to render novel views of the scene at arbitrary camera poses.

\subsection{Collision Detection}

Point clouds can be extracted from the trained NeRF model. This can be used as the input to the collision detection pipeline. Although It is possible to triangulate the point cloud into a mesh, it's challenging to further process it to be convex and enclosed, as required by simulation frameworks such as pybullet \cite{BulletRealTimePhysics2022} and MuJoCo \cite{MuJoCoAdvancedPhysics}. Therefore, A much simpler framework \textit{Flexible Collision Library} (FCL) (\cite{FlexiblecollisionlibraryFcl2024}) is used. FCL is a C++ library focused solely on detecting collision between objects. It supports mesh, octree or native shape primitives as object representations.

Collision detection using point cloud suffers from low efficiency. The point cloud can be converted into a list of occupied boxes (i.e. voxels) to accelerate detection. However, Octree offers a far more efficient alternative. Proposed in \cite{meagherOctreeEncodingNew1980a}, Octree is a tree data structure occupancy mapping. Each bounding box volume is equally subdivided into 8 children bonding boxes. This is done recursively until a certain depth is reached. The octree can be queried for collision in $O(\log n)$ time complexity while maintaining the same space complexity as voxels.

OctoMap \cite{hornung13auro} is used to convert the point cloud into an octree, so it can be imported by the FCL library. Pybind11 \cite{pybind11} is further used to bind the collision detection C++ API to Python in order to interface with nerfstudio and RL algorithms.

\section{Implementation: Pre-Processing Pipeline}

\begin{figure*}[!htp]
    \centering
    \includegraphics[width=0.85\textwidth]{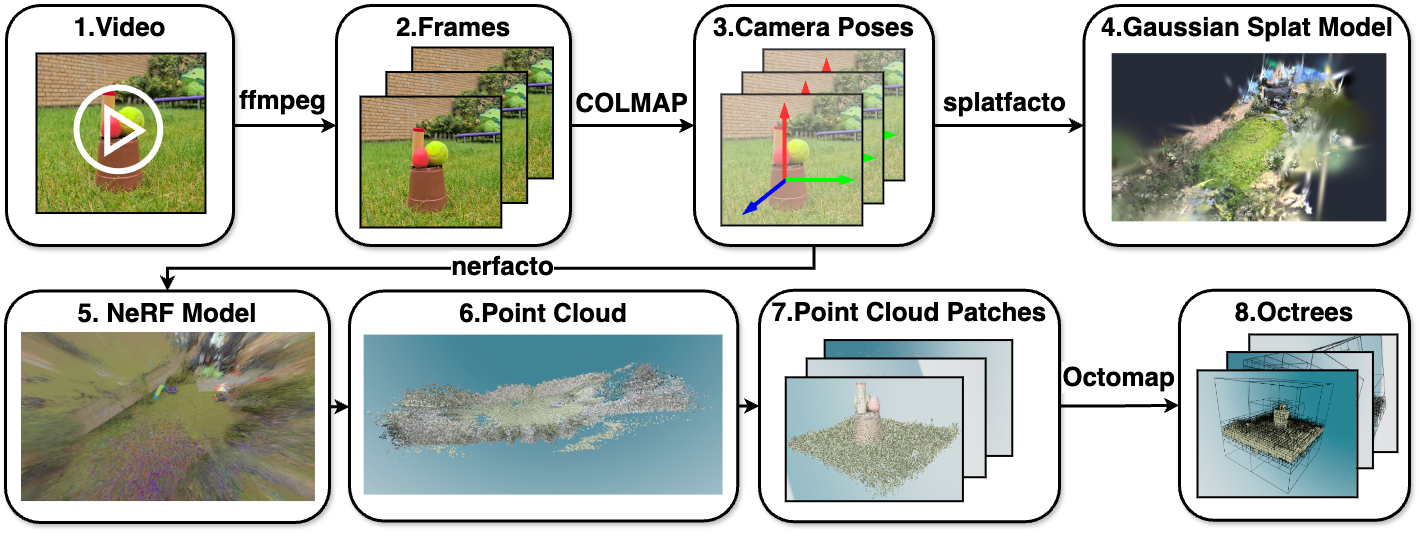}
    \caption{Pre-processing pipeline of the simulator.}
    \label{fig:pipeline}
\end{figure*}

From video to simulator, the processing pipeline is shown in Fig.~\ref{fig:pipeline}. Each component will be discussed in further detail in the following sections.

\subsection{Video Acquisition and Model Training}

Hand-held video is acquired using a mobile phone. The video is taken such that there is sufficient overlap of views and sufficient coverage of the scene, including the object of interest and the surrounding background. Frames are extracted from the video, then COLMAP \cite{schonbergerStructureMotionRevisited2016} is run to deduce the relative camera poses.nerfstudio is used to process and train a Gaussian Splatting model. The Model takes about 20 minutes to train on a 2070 Super GPU from a 1080p video.

\subsection{Extracting Point Clouds}

Due to the limitation of the nerfstudio package, point clouds can only be extracted from a NeRF model rather than a Gaussian Splatting model. Hence, a NeRF model is also trained using the same training data and the point cloud exported. The full point cloud contains artefacts and noise. Therefore, as the first step of the pipeline, the point cloud is cropped to remove spurious points at the edge of the scene. This is done by manually setting the x, y, and z limits of the point cloud.

\subsection{Octree Representation}\label{sec:octree}

The cropped point cloud is divided coarsely into voxels on a regular grid. Each voxel volume is then turned into an octree to represent fine details. Fig.~\ref*{fig:pipeline} step 7-8 shows an example of the transformation of a single voxel. As the point cloud is only used for collision detection, this downsampling does not sacrifice the visual fidelity of the simulator, while massively reducing the complexity of collision detection.

\begin{figure}
    \centering
    \includegraphics[width=0.3\textwidth]{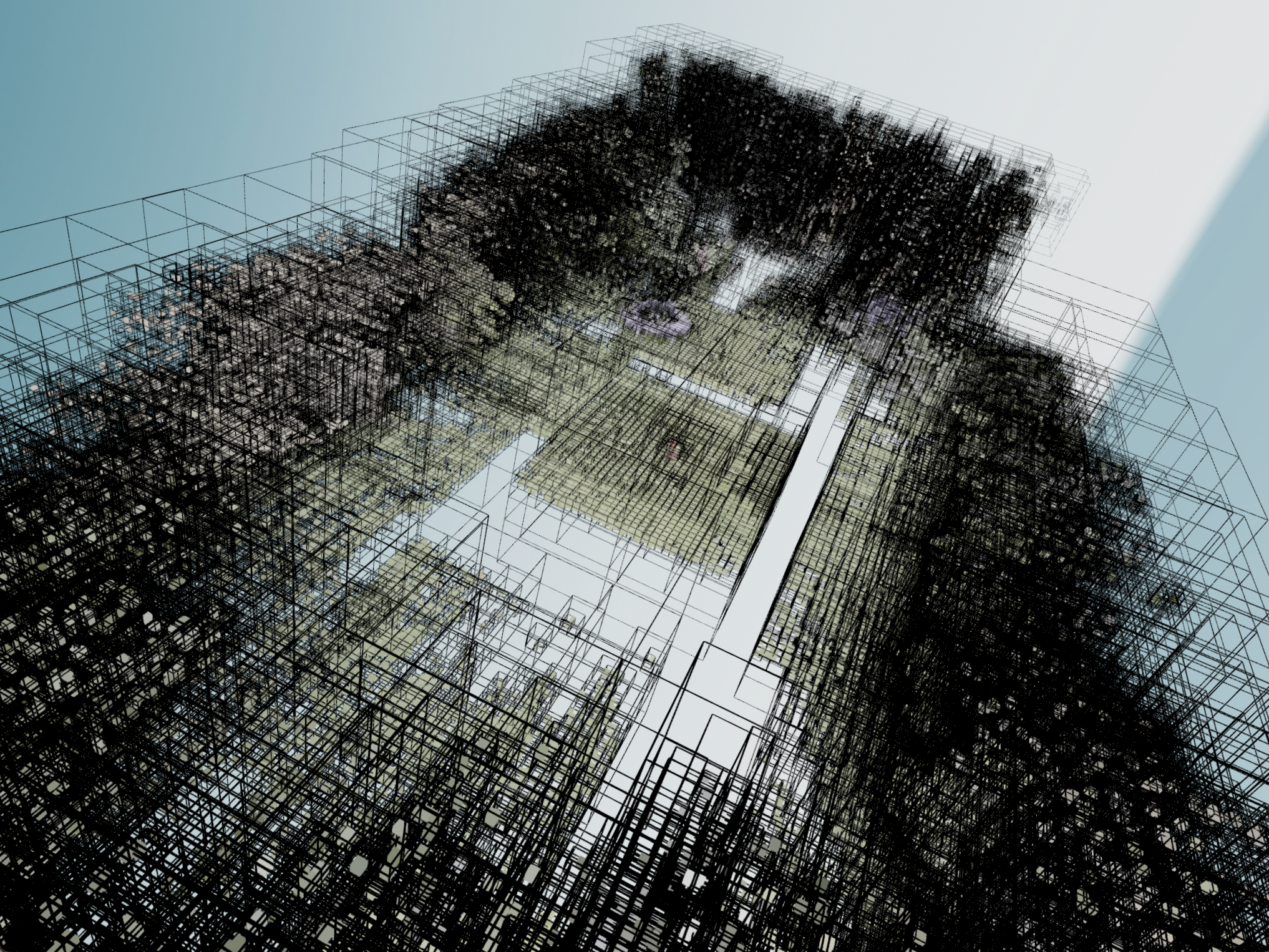}
    \caption{Combined octree representation of the garden scene}
    \label{fig:combined_octrees}
\end{figure}

The octree structure is 8 levels deep. This means the entire volume is divided into $64\time64\time64$ base voxels. The octree representation reduces the cropped point cloud shown in Fig.~\ref{fig:splatgym_simulator_overview.drawio} which contains $287,069$ points to $10,196$ occupied voxels. It takes less than $5\mu s$ to query the single tree for collision detection. A total of 75 octrees are constructed for the 75 voxels in the scene. Fig.~\ref{fig:combined_octrees} shows an illustration where all of them are put together to form the occupied space of the whole garden scene.

\section{Implementation: Software Architecture}

\begin{figure}
    \centering
    \includegraphics[width=0.49\textwidth]{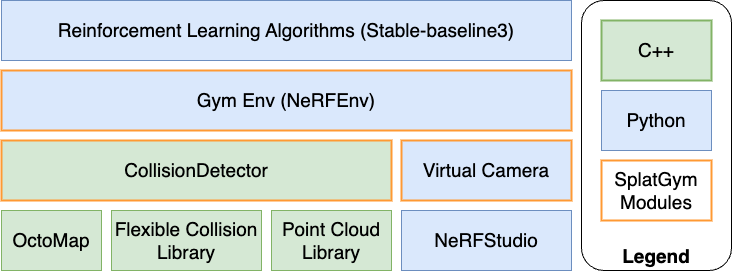}
    \caption{Software architecture of the simulator.}
    \label{fig:software_architecture}
\end{figure}

At runtime, the simulator software stack loads the Gaussian splatting model together with the point cloud and presents an interface suitable for reinforcement learning. The software architecture is shown in Fig.~\ref{fig:software_architecture}.

\subsection{Virtual Camera}

The simulator implements the concept of an ego camera which moves with 6DoF in the free space of the scene. An image can be extracted at any given camera pose. This is abstracted into a Python class. It exposes a simple API to move the camera and extract images at any resolution. It interacts with the Gaussian Splatting model in the back end to render the images.

\subsection{Collision Detector}

The Point Cloud file is loaded using the Point Cloud Library (PCL) then converted to an octree representation. A virtual bounding box is created to account for the space occupied by the ego camera. Flexible Collision Library (FCL) is used to detect collision between the camera bounding box and the underlying scene octree. OctoMap, PCL and FCL APIs are all in C++. All this functionality is integrated into a C++ class implementing a simple API. Python bindings for the C++ code allow collision detector to be called directly from high level abstractions. A single instance of collision detector manages a single octree. The simulator creates multiple instances of the collision detector for the whole scene.

\subsection{Gym Environment}

\begin{figure}
    \centering
    \includegraphics[width=0.19\textwidth]{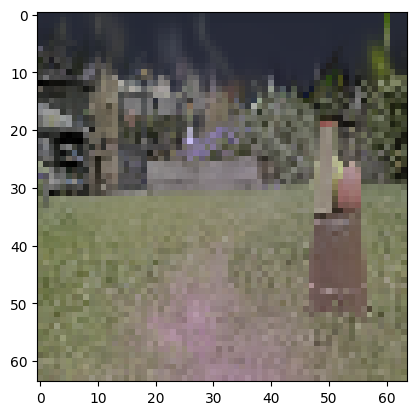}
    \includegraphics[width=0.29\textwidth]{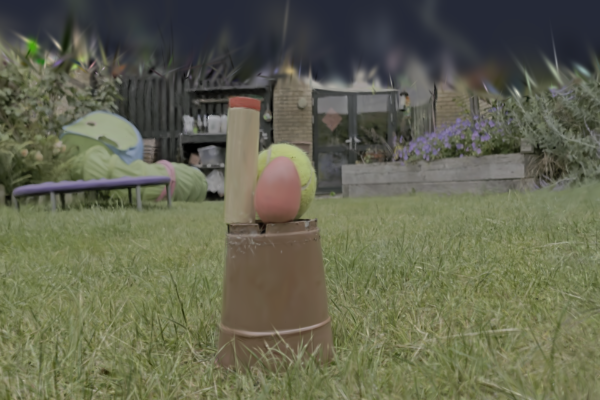}
    \caption{Example observation space of the robot (left) and a higher resolution render when the ego camera is at the goal position (right).}
    \label{fig:observation}
\end{figure}

To work with open-source RL libraries, the simulator implements the Gymnasium (\cite{towersGymnasium2024}) Env API. Gymnasium is a popular standard for specifying environments for reinforcement learning. Wrapping a Markov Decision Process (MDP), the API requires functions such as reset, step, and render to be implemented. To demonstrate the capability of the simulator for reinforcement learning, the following free space navigation problem is set up:

\begin{itemize}
    \item \textbf{Goal}: The robot finds the red egg in the scene. I.e., the robot can navigate to the pose corresponding to the image in Fig.~\ref{fig:observation}.
    \item \textbf{Action Space}: The robot can move a fixed-size step in the x or y direction, or rotate a fixed angle around the z-axis. This results in 3 degrees of freedom and 6 total actions.
    \item \textbf{State Space}: Corresponding to the action space, the robot can be in a finite number of locations in the x, and y planes as well as a finite number of orientations around the z-axis.
    \item \textbf{Observation Space}: The robot observes only the image from the ego camera. It is a 3-channel RGB image of size 64x64. The robot does not have direct access to its state information. Everything must be inferred from the ego image. An example is shown in Fig.~\ref{fig:observation}.
    \item \textbf{Reward}: The reward is sparse and is awarded according to the following criterion:
    \begin{itemize}
        \item +50 if the robot is at the goal position.
        \item -0.2 for each step taken. This is to encourage the robot to reach the goal as quickly as possible.
        \item -10 if the robot collides with the underlying scene. This allows the robot to learn to avoid obstacles.
        \item -10 if the robot is out of the [-1, 1] range in either the x or y direction. This limits the playing space so that the policy converges quickly.
    \end{itemize}
    \item \textbf{Termination}: The episode terminates when the robot reaches the goal, goes out of bounds, or collides with the underlying scene. There is also a maximum episode length of 200 steps to avoid wasting compute resources when the policy gets stuck in a loop.
\end{itemize}

\subsection{Curriculum Learning}

Curriculum learning is the idea to introduce training scenarios to the agent in a gradually increasing difficulty. This is to help the agent to learn faster and more robustly. This approach has been demonstrated to work well in robot control problems \cite{sangerNeuralNetworkLearning1994} and has been widely applied in modern RL problems \cite{JMLR:v21:20-212}.

Following this approach, an automatic Curriculum is implemented for the free space navigation problem in the SplatGym environment. At the beginning of learning, the robot is initialized close to the goal position. As the robot learns to reach the goal consistently, the initialization position is moved further away from the goal. As learning progresses, the initialization is moved further and further away until the robot can reach the goal from the edge of the play area. At this point, the Curriculum is finished, and the initial position is chosen randomly so that the robot can reinforce its learning.

\subsection{The Garden Scene}

A scene is captured of a garden. It consists of a grass section in the middle, surrounded by trees, brick walls and houses. Objects are placed in the middle of the grass section to serve as a target for navigation. This setting represents an unstructured outdoor natural environment common in agriculture, forestry or conservation robotic applications. The use of a red egg and apple as the target object mimic scenarios arising from fruit-picking applications. The garden scene is relatively large, but has fine-grained geometric detail around the target object. This necessitates the octree representation described in Section~\ref{sec:octree} to efficiently cover an expansive space while preserving fine details. The scene is also challenging visually due to the repeating patterns of the grass and trees, the monotonous green colour, and the occlusion of the target object by other objects. This makes it a good showcase for the simulator.

\section{Results \& Discussions}

\subsection{Experiment I: Free Space Navigation Problem}

\begin{figure}
    \centering
    \includegraphics[width=0.49\textwidth]{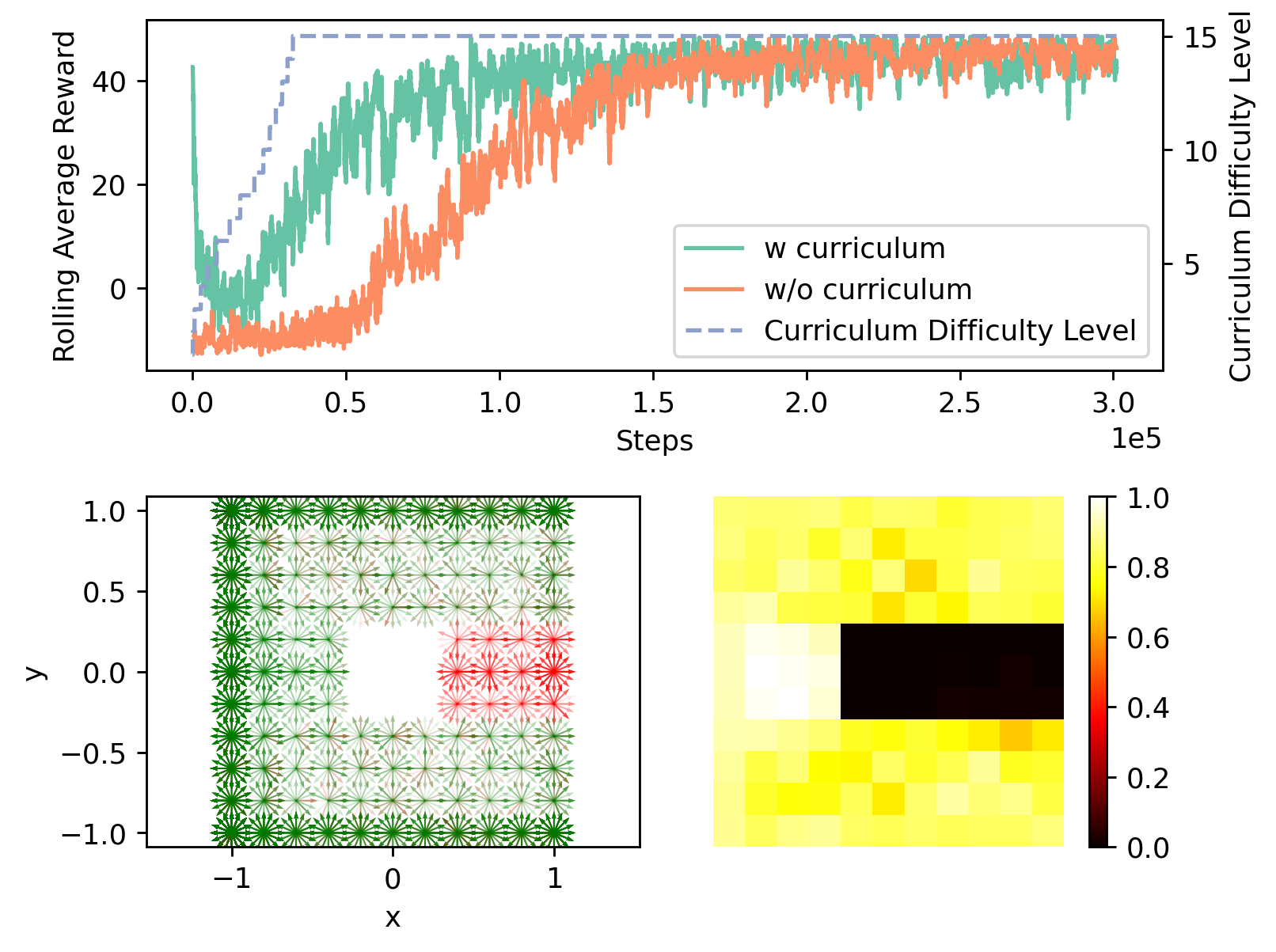}
    \caption{Result for Experiment I. \textbf{Top}: Rolling average reward during training. \textbf{Bottom left}: Starting positions and orientations of the robot
    for each episode during training. Green indicates a successful episode, red otherwise. Darker colours
    indicate the robot started in that pose more often. \textbf{Bottom right}: Success rate of the robot starting at different locations.}
    \label{fig:rolling_average_reward}
\end{figure}

Using the garden scene, a navigation policy is trained using PPO algorithm (\cite{schulmanProximalPolicyOptimization2017}) for $30,000$ steps. The goal of the robot is to find the red egg. The simulator and RL algorithm run at over 100Hz on a 2070 Super GPU. The rolling average reward is plotted in Fig.~\ref{fig:rolling_average_reward}. At the end of the training, the policy can consistently find the goal state, obtaining the final reward of 50. The success rate of the robot starting at different locations is shown in Fig.~\ref{fig:rolling_average_reward}. The success rate is high throughout the map, higher when closer to the goal pose at mid left-hand side of the plot. But the policy distinctly fails when starting at mid right-hand side of the plot. This corresponds to the location where the red egg is hidden behind other objects. The starting positions and orientations of the robot for each episode during the training are shown in Fig.~\ref{fig:rolling_average_reward}. A nice coverage of all the possible states is achieved throughout the training. The middle empty square corresponds to where the camera collides with the underlying scene.

Fig.~\ref{fig:rolling_average_reward} also shows the learning curve without the curriculum. In this case, the policy takes significantly longer to converge. Curriculum Learning helps the policy to find good state action pairs close to the goal at the beginning of the training. Thus allowing the policy to learn faster and more robustly.

\subsection{Experiment II: More complex state space}

The gym environment can be easily extended and customized for different problem setups. The NeRFFPSEnv class implements an extension of the free space navigation problem. The ego camera moves in the x, y plane and always travels forward and backward in its current heading. It's also allowed to rotate about the z-axis. In this instance, the state space is no longer on a regular grid and is much larger and more complex than Experiment I.

\begin{figure}
    \centering
    \includegraphics[width=0.49\textwidth]{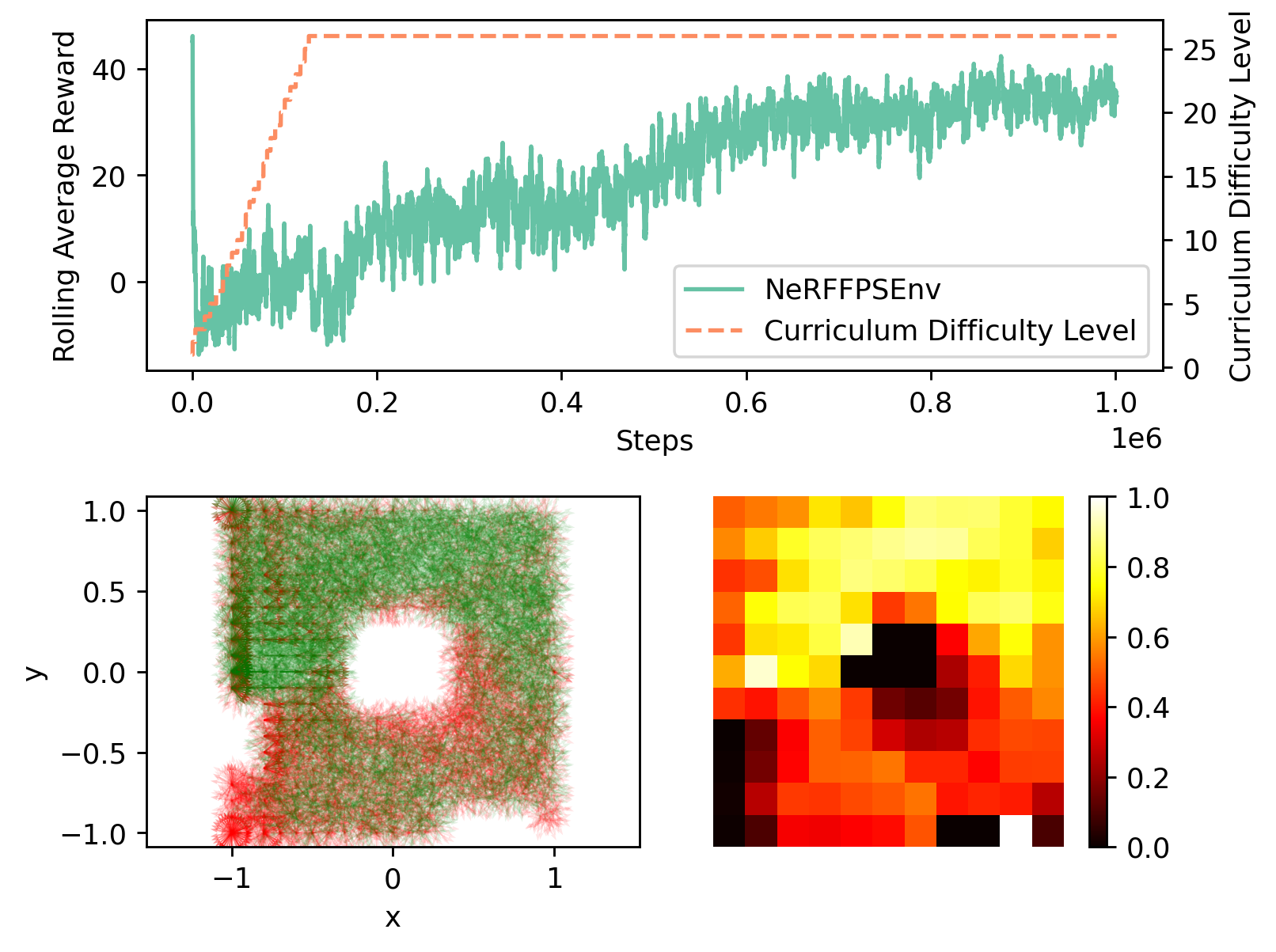}
    \caption{Result for Experiment II. \textbf{Top}: Rolling average reward during training with NeRFFPSEnv. \textbf{Bottom left}: Starting positions and orientations of the robot for each episode during training. \textbf{Bottom right}: Success rate of the robot starting at different locations.}
    \label{fig:fps_learning_curve}
\end{figure}

The learning curve is plotted in Fig.~\ref{fig:fps_learning_curve}. With the increased complexity of the state space, the policy takes longer to converge. Taking $10,000$ steps just to complete the Curriculum, compared to about $3,000$ steps for the simpler free space navigation problem. The policy did not fully converge even after a million steps of training. Notably, this setup of action and state space allows the policy to output much smoother and realistic robot movement, which will improve the usability of the policy in a real-world application.

\subsection{Experiment III: Overlaying Virtual Objects}

\begin{figure}
    \centering
    \includegraphics[width=0.49\textwidth]{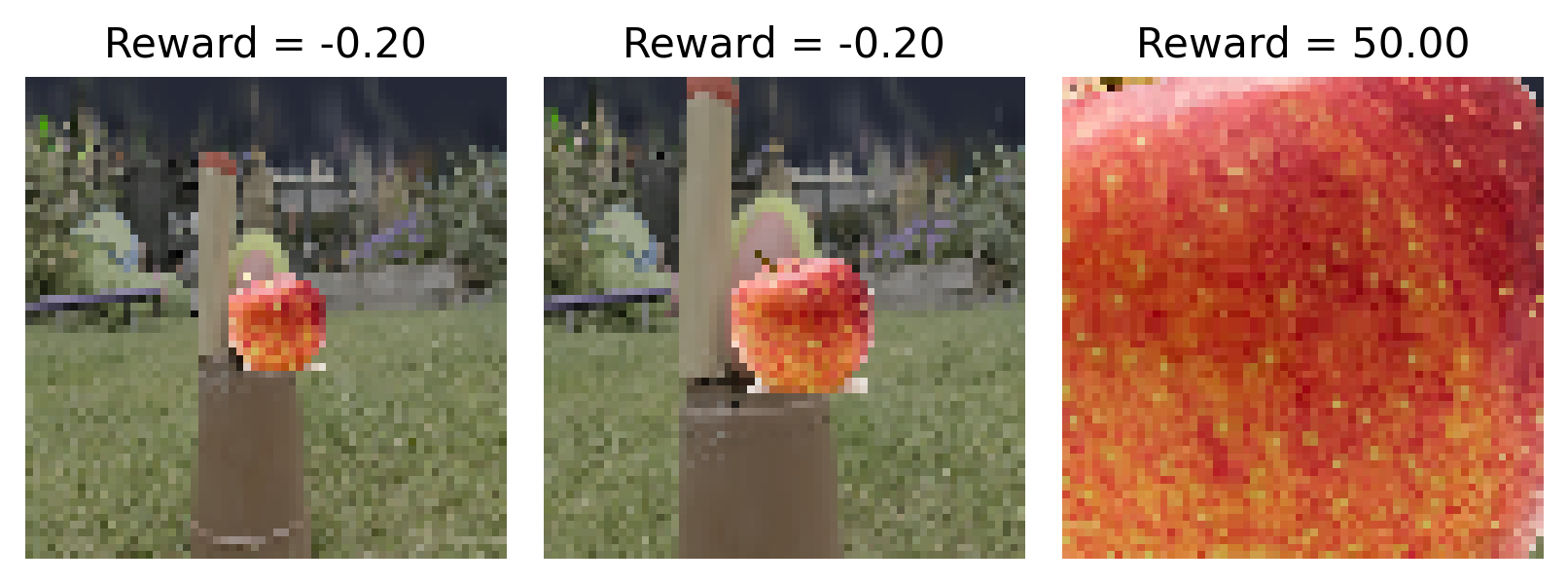}
    \caption{Overlaying virtual objects (apple) on top of the scene. The plots show the observation image of the robot as it is walking towards the apple. The robot obtains a large positive reward by moving close to the apple.}
    \label{fig:overlayed_objects}
 \end{figure}

Virtual objects can be added to the simulator. A static sprite of an apple is overlaid on top of the scene. It has a fixed 3D location and is projected dynamically into the image plane as the camera moves around. The sprite is then blended with the ego camera image. Fig.~\ref{fig:overlayed_objects} shows the observation of the robot as it is walking towards the apple. A large positive reward can be obtained when the observed image contains enough red pixels.

\begin{figure}
    \centering
    \begin{subfigure}[t]{0.24\textwidth}
        \includegraphics[width=\textwidth]{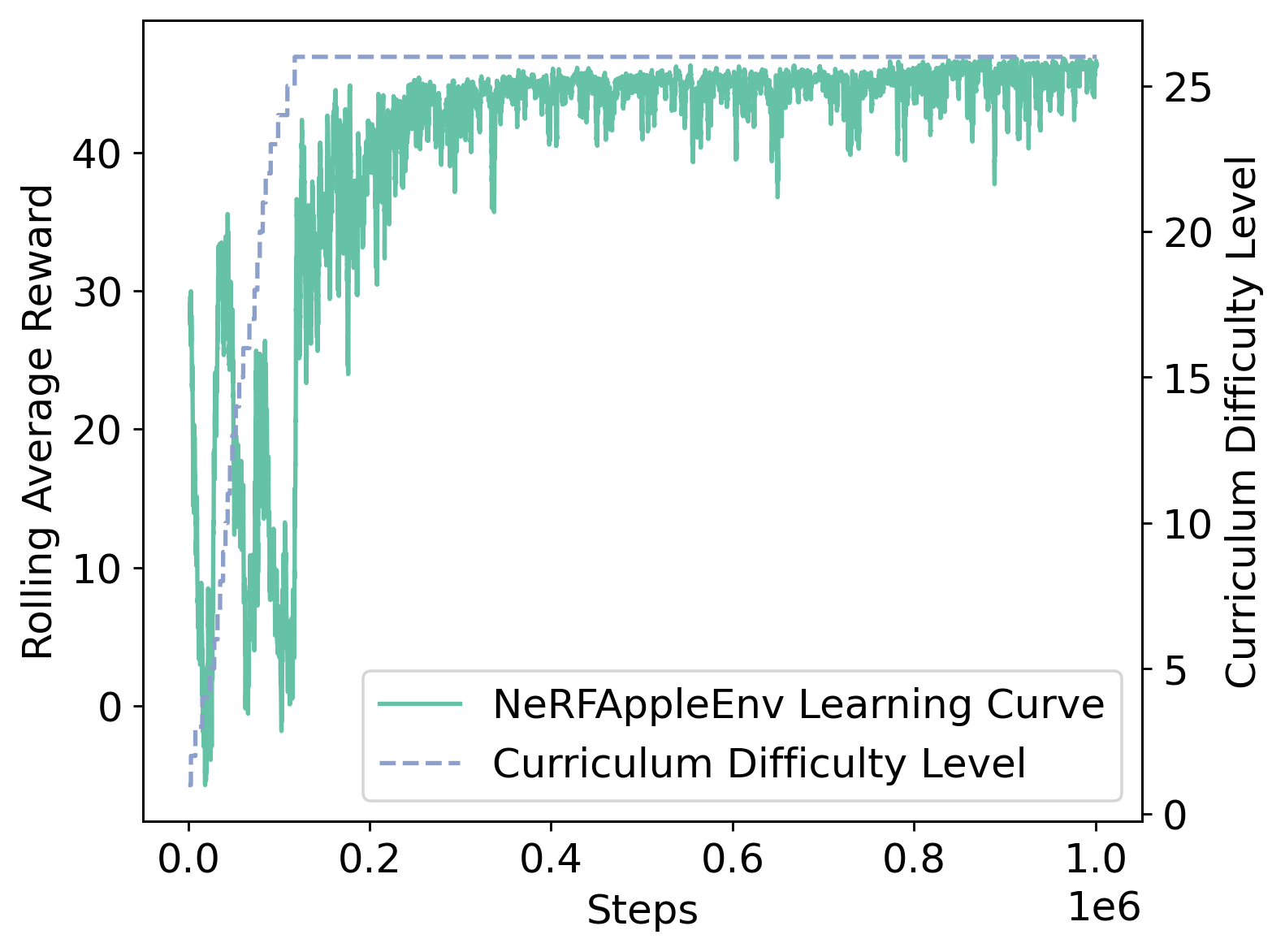}
        \caption{Rolling average reward obtained for episodes during training with the apple overlaid environment.}
        \label{fig:apple_learning_curve}
    \end{subfigure}
    \hfill
    \begin{subfigure}[t]{0.24\textwidth}
        \includegraphics[width=0.92\textwidth]{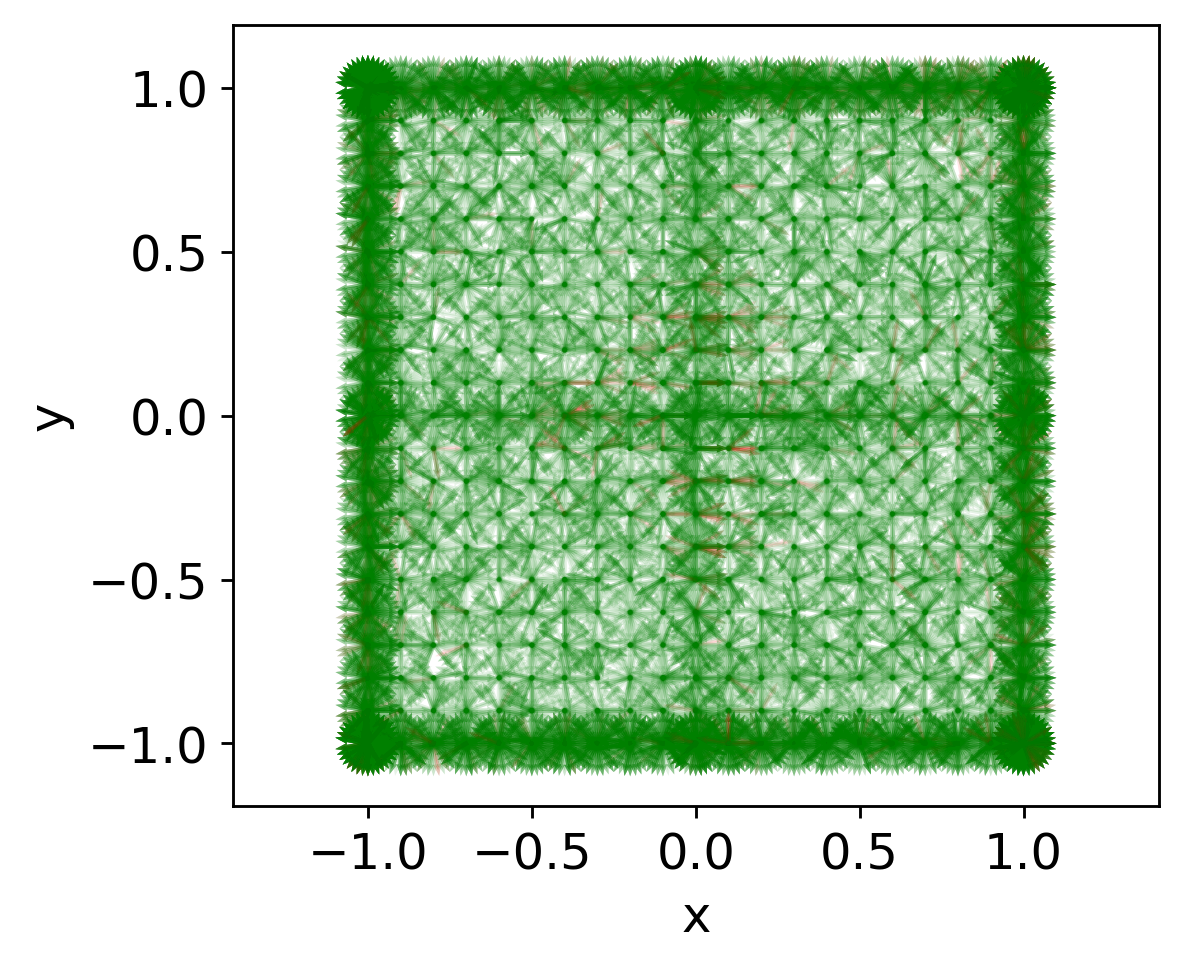}
        \caption{Starting positions and orientations for each episode during the training with the apple overlaid environment.}
        \label{fig:apple_starting_positions}
    \end{subfigure}

    \caption{Results from the apple learning environment.}
    \label{fig:apple_learning}
\end{figure}

A policy is trained using PPO with Curriculum Learning. The result shows that the robot can consistently walk towards the apple regardless of starting pose (Fig.~\ref{fig:apple_learning_curve}). A strategy emerged where the robot will spin around to look for the apple if the apple is not in view. Once apple is found the ego robot moves towards the apple, adjusting heading along the way.

\subsection{Experiment IV: Sim-to-real transfer}

Experiments are conducted to evaluate the transferability of the trained policy to the real world. A real apple is placed in the middle of the garden. This scene is videoed and processed into a simulation. A large reward is given for reaching the apples' location. A policy is trained following the same curriculum learning setup as before. Random noise is added to the camera pose to regularize the policy to be more robust to unsteady camera poses in the real world.

Test videos are taken of the same scene, with the cameras approaching the apple from different starting positions. A total of 6 clips are captured, covering a diverse range of viewpoints around the apple. Each frame of the test videos is manually labelled with an optimal action. The same frames are fed into the trained policy to obtain the predicted action. The rate at which the predicted action matches the optimal action is calculated. The results are shown in Tab.~\ref{tab:sim_to_real}.

Regularization consistently improves the real world performance of the policy. Across all test cases, the predicted action matches the manual labelled action in more than 78\% of the frames. In individual test cases, the similarity rate reach as high as 87.5\% while maintaining above 66\% across the board. This indicates that the trained policy exhibits navigational decision-making closely aligned with that of a human operator. This outcome provides further validation that the SplatGym environment is a viable platform for training policies capable of zero-shot transfer to real-world applications.

\begin{table}
    \centering
    \caption{Sim-to-real transfer results}
    \label{tab:sim_to_real}
    \begin{tabular}{p{1.05cm}p{0.6cm}p{0.6cm}p{0.6cm}p{0.6cm}p{0.6cm}p{0.6cm}p{0.6cm}}
        \toprule
        \textbf{Test Case} & \textbf{1} & \textbf{2} & \textbf{3} & \textbf{4} & \textbf{5} & \textbf{6} & \textbf{Total} \\
        \midrule
        \textbf{frame count} & 282 & 333 & 510 & 158 & 164 & 121 & 1568 \\
        \midrule
        \textbf{w/o reg*} & 0.759** & 0.408 & 0.690 & 0.797 & 0.451 & 0.479 & 0.612 \\
        \textbf{with reg} & 0.762 & 0.664 & 0.875 & 0.873 & 0.756 & 0.694 & 0.783 \\
        \bottomrule
    \end{tabular}
    \caption*{*Comparison between the policy trained with and without regularization.\newline**Rate at which the predicted action from the policy matches that of the human label.}
\end{table}

\section{Conclusion \& Future Work}

SplatGym fills a gap in existing research of an open-source general-purpose neural environment for robotic learning. By providing convenient and unrestricted tooling, and by eliminating the need for the verbose conventional 3D simulations, it enables deep reinforcement learning to be applied to a wider range of robotic problems.

Using a single video as input, the SplatGym generates photo-realistic images of the scene from any arbitrary camera pose and detect collision between the camera and the underlying scene objects. The simulator is implemented as a gym environment and works seamlessly with existing reinforcement learning libraries. Virtual objects can be in-painted into the camera view to enable more complex tasks.

To highlight the capabilities of the simulator, several navigation policies are trained using reinforcement learning. The policies are shown to have high rate of success in the simulation environment. The policies transfer to the real world zero-shot and is demonstrated to operate similar to a human.

The software stack is implemented in a modular way and can be easily extended to include more complex scenes and tasks. The simulator is fully open-source, and we look forward to collaboration with researchers in the field to further develop the functionalities. All documentation and code are available at \url{github.com/SplatLearn/SplatGym}.

The simulator has the potential to tackle a wide range of reinforcement learning problems. A few promising directions can be taken for future exploration.

\begin{enumerate}
    \item \textbf{Integrate the embodiment of the robot} - This requires more explicit modelling of the geometry and mechanics of the robot. The policy can then learn implicitly the forward and inverse kinematics of the robot.
    \item \textbf{Interaction with Scene Objects} - To simulate tasks such as fruit picking or pushing objects.
    \item \textbf{Learning from Demonstration} -  Adding recording and replaying functionality to the simulator can allow the robot to learn from human demonstrations.
\end{enumerate}


\printbibliography

\end{document}